\documentclass[10pt,twocolumn,letterpaper]{article}

\usepackage[pagenumbers]{cvpr} 
\usepackage{color}

\definecolor{cvprblue}{rgb}{0.21,0.49,0.74}
\usepackage[pagebackref,breaklinks,colorlinks,allcolors=cvprblue]{hyperref}

\def\paperID{6581} 
\def\confName{CVPR}
\def\confYear{2025}
\usepackage{multirow}
\usepackage{amsmath}
\DeclareMathOperator*{\argmax}{argmax}

\def\img{img}
\def\imgf{I}
\def\imgencoder{f_{\theta_{i}}}
\def\txt{txt}
\def\txtf{T}
\def\txtencoder{g_{\theta_{t}}}
\def\method{LVP-CLIP}
\def\methoda{\method-I}
\def\methodb{{\method}-IT}
\def\methodc{{\method}-C}
\def\labely{y}
\def\tasknum{M}
\def\samplenum{N}
\def\labelv{L}
\def\cbf{IT}
\def\dim{\mathbb{R}^{D}}
\def\ebdim{D}
\def\ebf{E}
\def\poolsize{P}
\def\labelpool{LVP}
\def\calnum{O}
\def\clsnum{K}
\def\cls{k}

\title{\method: Revisiting CLIP for Continual Learning with Label Vector Pool}
\author{Yue Ma \\ 
Syracuse University\\
{\tt\small yma183@syr.edu}
\and
Huantao Ren \\ 
Syracuse University\\
{\tt\small hren11@syr.edu}
\and
Boyu Wang \\ 
Syracuse University\\
{\tt\small bwang30@syr.edu}
\and
Jingang Jin \\ 
Syracuse University\\
{\tt\small jjin24@syr.edu}
\and
Senem Velipasalar \\ 
Syracuse University\\
{\tt\small svelipas@syr.edu}
\and
Qinru Qiu \\ 
Syracuse University\\
{\tt\small qiqiu@syr.edu}
}

\begin{document}
\maketitle
\begin{abstract}
Continual learning aims to update a model so that it can sequentially learn new tasks without forgetting previously acquired knowledge. Recent continual learning approaches often leverage the vision-language model CLIP for its high-dimensional feature space and cross-modality feature matching. 
Traditional CLIP-based classification methods identify the most similar text label for a test image by comparing their embeddings. However, these methods are sensitive to the quality of text phrases and less effective for classes lacking meaningful text labels. In this work, we rethink CLIP-based continual learning and introduce the concept of Label Vector Pool ({\labelpool}). {\labelpool} replaces text labels with training images as similarity references, eliminating the need for ideal text descriptions. We present three variations of {\labelpool} and evaluate their performance on class- and domain-incremental learning tasks. Leveraging CLIP's high dimensional feature space, {\labelpool} learning algorithms are task-order invariant.~The new knowledge does not modify the old knowledge, hence, there is minimum forgetting. Different tasks can be learned independently and in parallel with low computational and memory demands. 
Experimental results show that proposed {\labelpool}-based methods outperform the current state-of-the-art baseline by a significant margin of 40.7\%. 
\end{abstract}

\section{Introduction}
\label{sec:1}

\begin{figure}[t]
    \centering
    \includegraphics[width=1\linewidth]{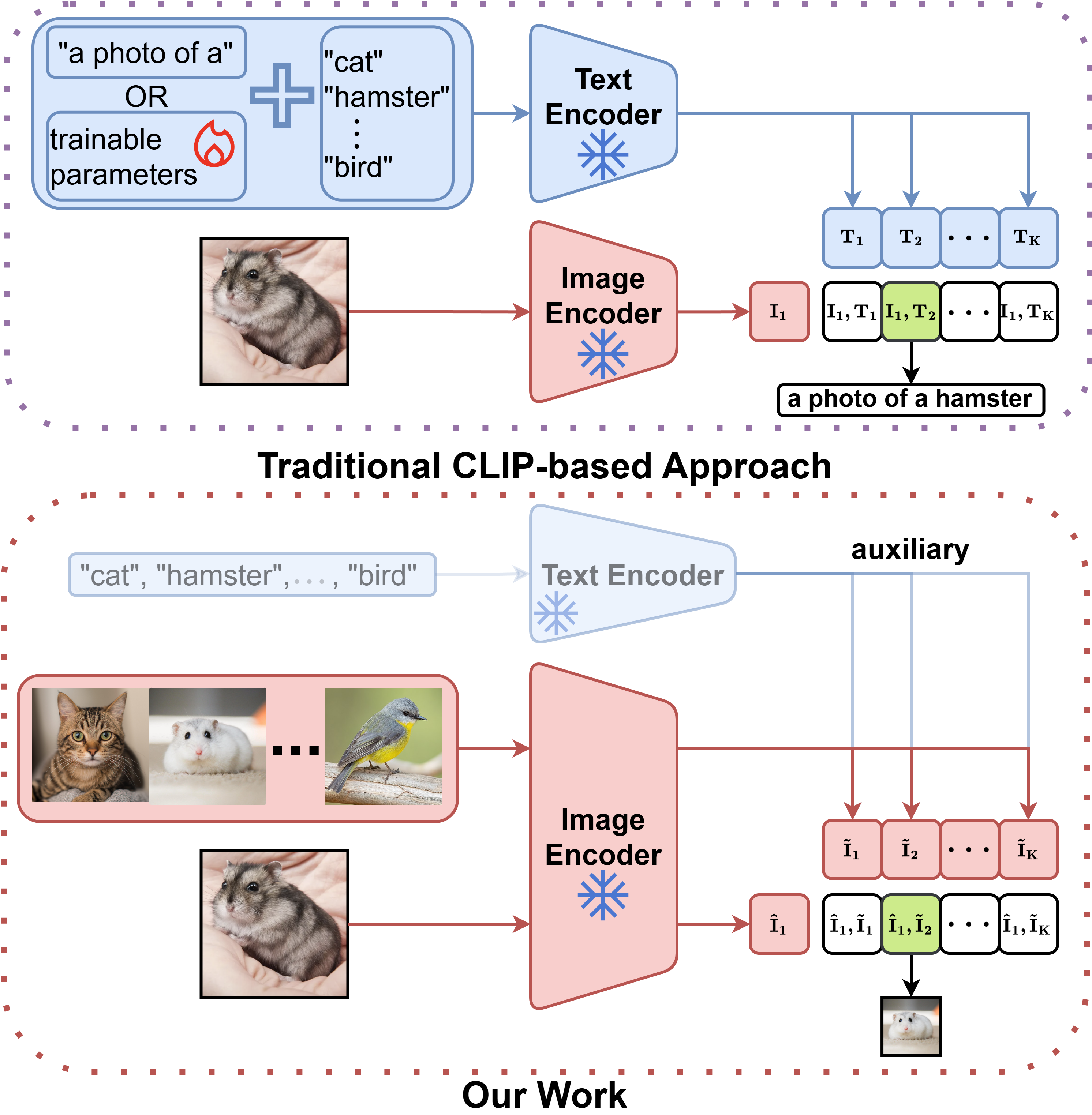}
    \vspace{-0.5cm}
    \caption{Comparison with traditional CLIP-based approaches. While traditional methods compare similarity between the encoded test image and text labels, our approach evaluates similarity between image embeddings directly and makes the text encoder play an auxiliary role when possible. }
\label{fig:intro}
 \vspace{-0.2cm}
\end{figure}

Deep neural networks trained by using supervised learning have achieved remarkable accuracy in classification tasks. Their effectiveness relies on the assumption that the training data distribution fully and accurately represents the testing data. However, in real-world scenarios, data samples are often not available all at once. New classes of knowledge are discovered sequentially and corresponding training data arrives in stages. Continual learning addresses this by incrementally training a model to effectively learn new tasks without catastrophic forgetting~\cite{Catastrophic_Interference_in_Connectionist_Networks} of previously acquired knowledge. Recent continual learning approaches frequently utilize the vision-language model CLIP (Contrastive Language-Image Pretraining)~\cite{CLIP} due to its high-dimensional feature space and ability to match features across different modalities.


CLIP bridges the gap between language and vision through contrastive learning, enabling zero-shot classification by matching a list of text embeddings to the test image embedding. The primary challenge of CLIP lies in the vast search space for the optimal text, since different text inputs can yield highly different results. Recent research works have focused on improving the text embedding quality. For example, PointClip~\cite{pointclip} leverages human knowledge and experience to design text descriptions for point cloud data. PointClipV2~\cite{pointclipV2} uses language models, like GPT-3~\cite{GPT3}, to generate enhanced text. Methods like CoOp~\cite{CoOp} and CoCoOp~\cite{CoCoOp} further improve performance by incorporating trainable parameters into the text encoder. Yet, all these approaches still classify the input image by searching for the best matching text embedding.

L2P~\cite{L2P} first introduces the prompt-pool on pre-trained ViT~\cite{ViT} for continual learning, leveraging trainable prompts inside a pool to retain knowledge from different tasks. Following works~\cite{DualPrompt,S-Prompts,CODA-Prompt,boyu_ecai} improve L2P by either redesigning the prompts and prompt pools or refining the matching procedure between the prompts and embeddings. A challenge for prompt-pool-based methods is the uncertainty in the matching procedure between the test embedding and the prompt key. Although various optimization objectives are designed to improve matching, there is no guarantee that the optimal prompt will be selected for a given test image.~If an incorrect prompt key is chosen, leading to the selection of wrong prompts, the result is more likely to be inaccurate, since the prompt may introduce biased information instead of providing helpful knowledge. Furthermore, as the same set of prompts is updated across different tasks, forgetting is unavoidable.

In this work, we revisit image classification and continual learning by introducing a novel concept, referred to as the Label Vector Pool ({\labelpool}).
As shown in \cref{fig:intro}, instead of searching for the optimal text label phrases and relying only on their embeddings for similarity comparison, we utilize the image embeddings generated from training images as references.
The proposed approach, utilizing LVP, enables CLIP-based incremental learning models to reduce if not remove their dependency on the quality of label phrases.
{\labelpool} allows CLIP to adapt to a wide range of datasets, particularly those with classes difficult to describe in text or class names that have no semantic meaning, such as ``ZIL103" or ``ZSU234", etc. Furthermore, feature vectors from the same modality tend to cluster more closely than those from different modalities, leading to improved classification accuracy.

Since we use CLIP as the foundation model and leverage its  embedding capabilities, we refer 
to our method as LVP-CLIP, which is a completely different approach compared to prompt-pool-based methods. Instead of constructing a prompt pool that provides additional features for the classifier, we consolidate the knowledge of each class into a single vector and store it directly in the pool. This single vector can be formed from training image embeddings or as a combination of image and text embeddings. Given a test image, we simply calculate the similarity between its embedding and each vector in {\labelpool}, then select the class with the highest similarity. 

Most existing class-incremental and domain-incremental learning algorithms assume that the upper bound of class count is known in advance. This assumption is necessary since those methods typically use an MLP-based classifier, where the number of classes determines the output layer size. When the number of categories exceeds the classifier’s predefined output dimension, the MLP must be expanded to accommodate the larger output and the entire model must be fine-tuned with data from all previously learned tasks. While a low upper bound results in frequent adjustments to the MLP size and parameter fine-tuning, a high upper bound leads to over provisioning. 
In contrast, our proposed {\method} imposes no restriction on the number of classes. Its memory cost grows linearly at a very low rate as the number of classes increases. In fact, we demonstrate its scalability by applying it to a cross-dataset, cross-domain incremental learning task, Cross-Task Incremental Learning (CTIL), which includes 595 classes, as discussed in detail in Sec.~\ref{sec:exp}.

{\method} avoids performance degradation as the number of tasks increases, since learning new information does not modify previously stored knowledge. In other words, it treats incremental learning and batch learning equivalently. As a similarity-based approach, it simply searches for the embedding vector in the feature space that is most similar to the input vector. The high dimensionality of its feature space provides substantial memory capacity. 
The main contributions of this work include the following:

\begin{itemize}
\item We propose the concept of Label Vector Pool (\labelpool) by revisiting the classification procedure of CLIP.  
As a similarity based approach like CLIP, the {\labelpool} uses image embedding or a mixture of image and text embeddings as references, hence is more flexible and less biased to any one modality compared to CLIP. 

\item We present three variations of {\labelpool}, namely {\methoda}, {\methodb} and {\methodc}. We evaluate our methods in class-incremental, domain-incremental and cross-task incremental settings and outperform SOTA methods.
\item The proposed {\labelpool}-based incremental learning has orders of magnitude lower computational complexity in learning and 2x lower inference complexity compared to baselines.
\item The performance of {\labelpool}-based learning is scalable to large number of tasks. We demonstrate that {\labelpool} has outstanding memory capacity and learning capability using an experimental setting that consists of 4 commonly used incremental learning datasets with 595 classes. 
\end{itemize}




\section{Related Work}
\label{sec:2}
{\bf Conventional solutions for continual learning.} 
Existing continual learning algorithms can be classified into three categories, namely regularization-based, architecture-based, and rehearsal-based methods.
Regularization-based methods~\cite{Memory_Aware_Synapses,Learning_without_Forgetting,Continual_Learning_Through_Synaptic_Intelligence,Overcoming_catastrophic_forgetting_in_neural_networks,Continual_Learning_of_a_Mixed_Sequence_of_Similar_and_Dissimilar_Tasks,Dreaming_to_Distill} set constraints on the trainable parameters by limiting the learning rate of the important parameters for old tasks.~Although these methods do not require additional memory for replay buffers and additional model parameters, they have limited performance on complex datasets.~Architecture-based methods~\cite{Learn_to_Grow,DER,A_Model_or_603_Exemplars,Deep_Bayesian_Unsupervised_Lifelong_Learning,Learn-Prune-Share_for_Lifelong_Learning,Overcoming_Catastrophic_Forgetting_with_Hard_Attention_to_the_Task,PackNet} create separate parameters for each task to bypass catastrophic forgetting. 
Rehearsal-based methods~\cite{Dark_Experience_for_General_Continual_Learning,Co2L,Online_Continual_Learning_with_Maximal_Interfered_Retrieval,Gradient_based_sample_selection_for_online_continual_learning,Large_Scale_Incremental_Learning,Experience_Replay_for_Continual_Learning} maintain a buffer of data from past tasks. During the optimization for new tasks, the model is also trained on buffered data from previous tasks to mitigate forgetting. The performance of these methods is limited by the buffer size, and as buffer size decreases the performance declines sharply. Additionally, these methods are challenging to apply when data is private or cannot be stored~\cite{Privacy_Preserving_Deep_Learning}. Our LVP-CLIP addresses continual learning challenge by extracting class knowledge directly from CLIP without relying on a rehearsal buffer.

\noindent {\bf Prompt-based continual learning.} Almost all recent works that achieved noteworthy performance in continual learning use a prompt-based architecture. After L2P~\cite{L2P} first proposed visual prompting, which constructs a prompt pool for continual learning tasks, prompt-pool-based methods have become the main track for its great performance. DualPrompt~\cite{DualPrompt} improves the pool design with task-invariant prompts (G-Prompt) for general knowledge and task-specific prompts (E-Prompt) for expert knowledge. S-Prompts~\cite{S-Prompts} designs domain-specific prompts and utilizes a K-NN operation as a domain identifier for inference. AttriClip~\cite{Attriclip} adapts the prompt pool for CLIP by maintaining a prompt pool for the text encoder. The prompt-pool-based methods may suffer from the mismatching challenge, since there is a potential to select the wrong prompt during inference. In addition, prompts add additional computation complexity in training and inference. These challenges do not apply to LVP. All of these recent continual learning works rely on the pre-trained models, such as ViT~\cite{ViT} and CLIP~\cite{CLIP}. Hence, we adopt the same foundational model for feature extraction and embedding.
\section{Preliminaries}
\label{sec:prelim}

For clarity, we use $\thicksim$ and $\wedge$ to differentiate the symbols for training and testing sets, respectively, in the rest of the paper. We use the superscripts to denote the index of classes or tasks and the subscripts to denote the index of training/testing instances.

\subsection{Continual Learning}

We consider a sequence of tasks $S=\{S^1,\cdots, S^\tasknum\}$, where $\tasknum$ is the total number of tasks.~Each task is a set, $S^t = \{(x_1^t,y_1^t),\cdots,(x_{n^t}^t,y_{n^t}^t)\}, t\in [1,...,\tasknum]$, where $(x_i^t,y_i^t), i\in[1,...,n^t]$, is a pair containing the input $x_i^t \in X$ and its corresponding label $y_i^t \in Y$ and $n^t$ is the total numbers of samples. The goal of continual learning is to train a model $f(\theta): X \rightarrow Y$ continuously over time on a set of tasks, arriving sequentially, such that it learns the new tasks without forgetting the old tasks.

Task-, Domain- and Class-Incremental Learning (TIL, DIL, CIL) are the three main scenarios of continual learning. 
Domain-incremental learning assumes that the number and labels of classes remain consistent across tasks, and the only difference among tasks is the distribution of the input data. Both Task- and Class-incremental learning address the scenario, where each task introduces a distinct set of new classes to be learned. Task-incremental learning assumes a known task identity at inference time, whereas the class-incremental learning does not make such assumption.  

\subsection{CLIP} \label{ssec:CLIP}

CLIP is a vision-language model trained on text-image pairs. It consists of a text encoder $\txtencoder(\cdot)$ and an image encoder $\imgencoder(\cdot)$. Given a sentence $\txt$ and an image $img\in \mathbb{R}^{H\times W\times C}$, the text and image embeddings are $ \txtencoder(\txt) = \txtf$ and   $\imgencoder(\img) = \imgf$  respectively, where $\txtf, \imgf\in \mathbb{R}^{D}$
with $D$ denoting the dimension of the embeddings. Given a dataset containing $\clsnum$ classes, the $\txt$ is a phrase like ``a photo of a [$\labely$]", where $\labely \in [1,...,\clsnum]$ is the index of the label, and [$y$] denotes the class name of the label. The probability of labeling the test image $\img$ with the class $\labely$ is computed as:
\begin{equation}
\small
  p(\labely|\img) = \frac{\exp{(\langle \txtf^{y},\imgf\rangle)}}{\sum^{\clsnum}_{\cls=1} \exp{(\langle \txtf^{\cls},\imgf \rangle)}},
\label{eq:clip1}
\end{equation}
where $\langle\cdot,\cdot\rangle$ denotes the similarity function.~Three commonly used similarity functions are L1 norm, L2 norm and cosine similarity, denoted by  $\langle\cdot,\cdot\rangle_{L1},\langle\cdot,\cdot\rangle_{L2},\langle\cdot,\cdot\rangle_{Cos}$, respectively. To classify an image, the class with the highest probability is chosen, i.e.
\begin{equation}
\small
\hat{y} = \argmax_{\cls} p(\cls|\img), \,\,\,\,\cls \in [1,...,\clsnum].
\label{eq:cliparg}
\end{equation}
\section{Methodology}
\label{sec:method}
In this section, we first introduce a superset method, {\labelpool}, by rethinking how CLIP can be used for classification. Then, we design three continual learning methods utilizing {\labelpool}, namely LVP-I, LVP-IT and LVP-C. Finally, we describe the 
loss functions to train these methods.

\subsection{Label Vector Pool(\labelpool)}
\label{sec:\labelpool}
As discussed in Sec.~\ref{ssec:CLIP}, CLIP classifies an input image by comparing the distance between the image embedding and a list of text embeddings that represent class labels.~The 
query vector with unknown-ID is compared with a group of key vectors with known IDs, and the query vector is assigned the ID of the key vector with the highest similarity. Following this concept, we define label/labeled vector $\labelv \in \dim$ as a vector with known ID, such as the class or domain name, or task ID, etc. A {\labelpool} for class $\cls$ is a set of label vectors with ID $\cls$, denoted as $\labelv^\cls = \{\labelv^\cls_1,\labelv^\cls_2,\cdots,\labelv^\cls_{\poolsize^\cls}\}, \text{where  }  \labelv^\cls_i\in\dim,i\in[1,..,\poolsize^\cls],\cls\in[1,...,\clsnum]$, and $\poolsize$ is the pool size. The similarity between an image embedding $I$ and $\labelv^\cls$ is computed as the maximum similarity between $I$ and all instances in $\labelv^\cls$:
\begin{equation}
\langle \labelv^{\cls},\imgf\rangle = \max{({\langle \labelv^{\cls}_1,\imgf\rangle},{\langle \labelv^{\cls}_2,\imgf\rangle},\cdots,{\langle \labelv^{\cls}_{\poolsize^\cls},\imgf\rangle})}.\\
\label{eq:label-pool}
\end{equation}
The probability that a testing image $\img$ belongs to class $y$ can be computed as:
\begin{equation}
\small
  p(\labely|\img) = \frac{\exp{(\langle \labelv^{y},\imgf\rangle})}
  {\sum^{\clsnum}_{\cls=1} \exp{(\langle \labelv^{\cls},\imgf \rangle)}}.
\label{eq:clip2}
\end{equation}
As can be seen, \cref{eq:clip1} is a special case of \cref{eq:clip2}, where the {\labelpool} contains only one labeled vector, which is the text embedding of the class name, i.e.,  $L^\cls = \{T^\cls\},\text{and } \poolsize^\cls=1$. We denote the computational complexity for classifying one image as $\calnum$,  and represent it using the number of times the similarity function is calculated, i.e., $\calnum =\sum_{\cls=1}^{\clsnum}{\poolsize^\cls}$. If all classes have the same pool size $\poolsize$, then the complexity simplifies to $\calnum =\sum_{\cls=1}^{\clsnum}{\poolsize^\cls} = \poolsize \times \clsnum$. 

Our {\labelpool}-based approach is a general framework for similarity-based classification and extends CLIP in two ways: (1) the test image is no longer restricted to comparison with text embeddings, it can be compared with any labeled vectors with matching dimensions; 
(2) each class can be represented by one or multiple labeled vectors, which may be obtained from different modalities. 

\begin{figure}[t]
    \centering
    \includegraphics[width=1\linewidth]{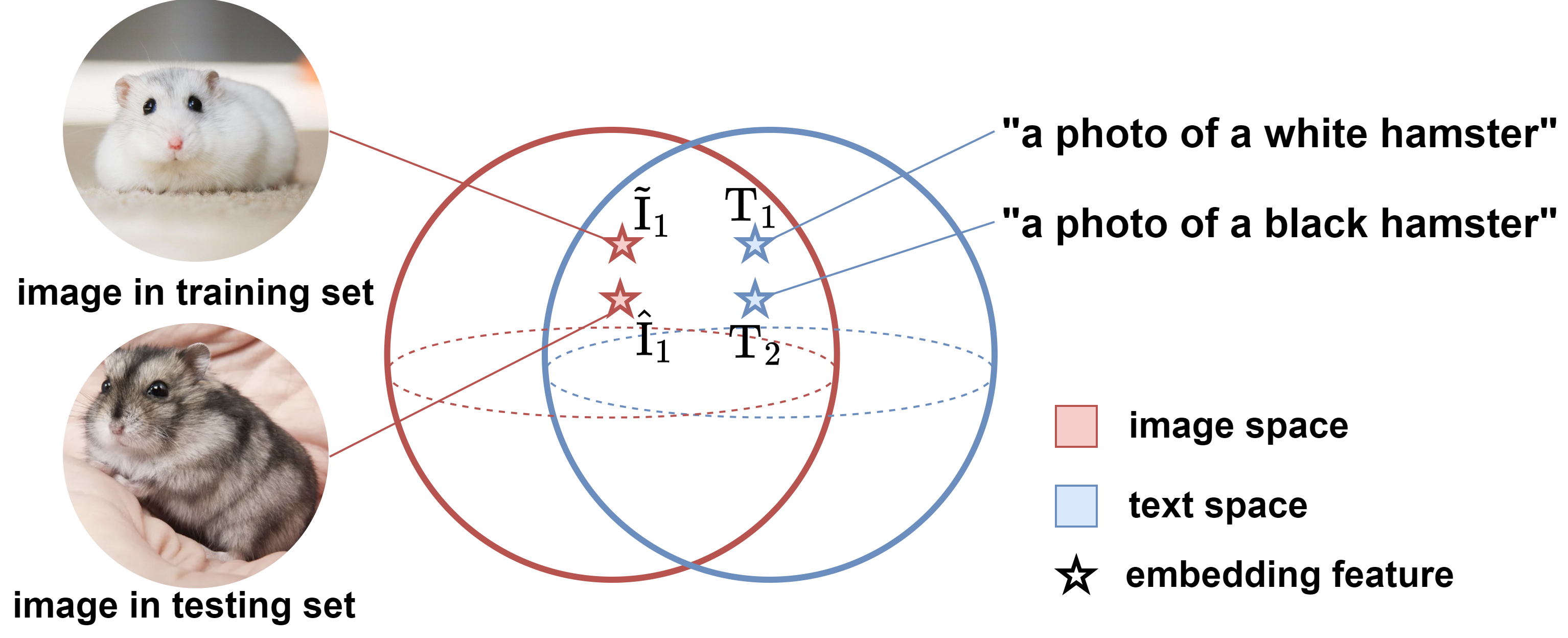}
    \caption{The hypothesis is that embeddings in the same modality should be more similar to each other.~The training image embedding $\Tilde{\imgf}_1$ is expected to be more similar to the test image embedding $\hat{\imgf}_1$ than the text embeddings $T_1$ and $T_2$, i.e., $\langle \hat{\imgf}_1,\Tilde{\imgf}_1\rangle > \langle \hat{\imgf}_1,\txtf_1\rangle, \langle\hat{\imgf}_1,\Tilde{\imgf}_1\rangle > \langle \hat{\imgf}_1,\txtf_2\rangle$.
    }
\label{fig:text-image}
\end{figure}

\subsection{Motivation for Using Image {\labelpool}}
The training of CLIP ensures that matching images and text phrases will be mapped to nearby locations in the feature space. Intuitively, we expect that embeddings of inputs from the same modality (e.g., image-to-image or text-to-text) will be closer to each other than those from different modalities, as illustrated in \cref{fig:text-image}.~As discussed in \cref{sec:\labelpool}, the task of classification is to identify the labeled vector most similar to the query. This raises an interesting question: can we use image embeddings by themselves in the {\labelpool} instead of text embeddings?
\begin{table}[th!]
  \centering
  \resizebox{0.85\linewidth}{!}{
  \begin{tabular}{cc cc cc cc cc cc cc}
    \toprule
     & 10\% & 30\% & 50\% & 70\% & 100\% &text\\
    \midrule
    \poolsize & 50   & 150  & 250  & 350   & 500   & 1 \\
    \calnum      & 5000   & 15000 & 25000 & 35000 & 50000 & 100  \\
    Acc.              & 71.0   & 74.8 & 76.1 & 76.9 & 78.2 & 73.3  \\
    \bottomrule
  \end{tabular}}
  \caption{Testing accuracy using embeddings of the training images (columns 2-6) and label text (column 7) as the {\labelpool} on CIFAR100. The percentage values in the first row represent the proportion of the training set included in the {\labelpool}. The second (P) and third (O) rows show the corresponding LVP size and the computational complexity of testing a single image.} 
  \label{tab:image-label}
\end{table}

\begin{figure*}
  \centering
   \includegraphics[width=1\linewidth]{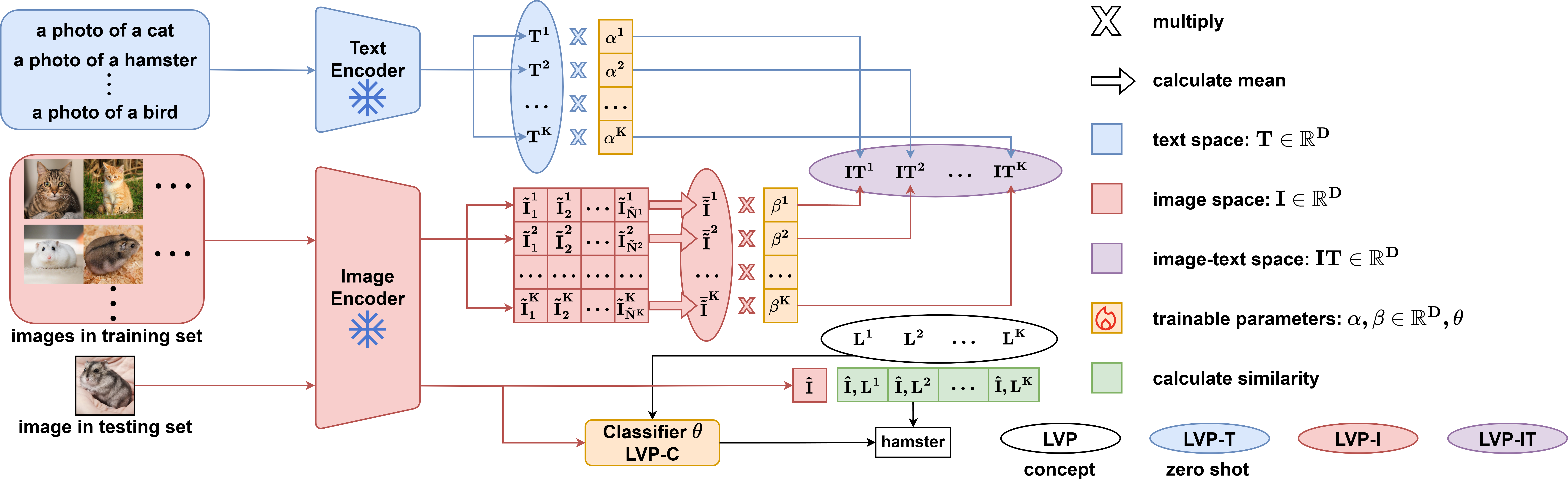}
   \caption{Framework of \method. Firstly, the concept of LVP is demonstrated. Secondly, three realizations of LVP is shown as LVP-T, LVP-I and LVP-IT. LVP-T known as zero-shot is LVP generated form the text encoder. Our proposed LVP-I is the mean of image embeddings of each class in the training set. LVP-IT can be obtained as a combination of LVP-T and LVP-I with the task-specific trainable paremeters $\alpha,\beta$ of each class. In addition, LVP-C is a classifier optimized on LVP.}
\label{fig:method}
\end{figure*}

\begin{figure}
  \centering
   \includegraphics[width=1\linewidth]{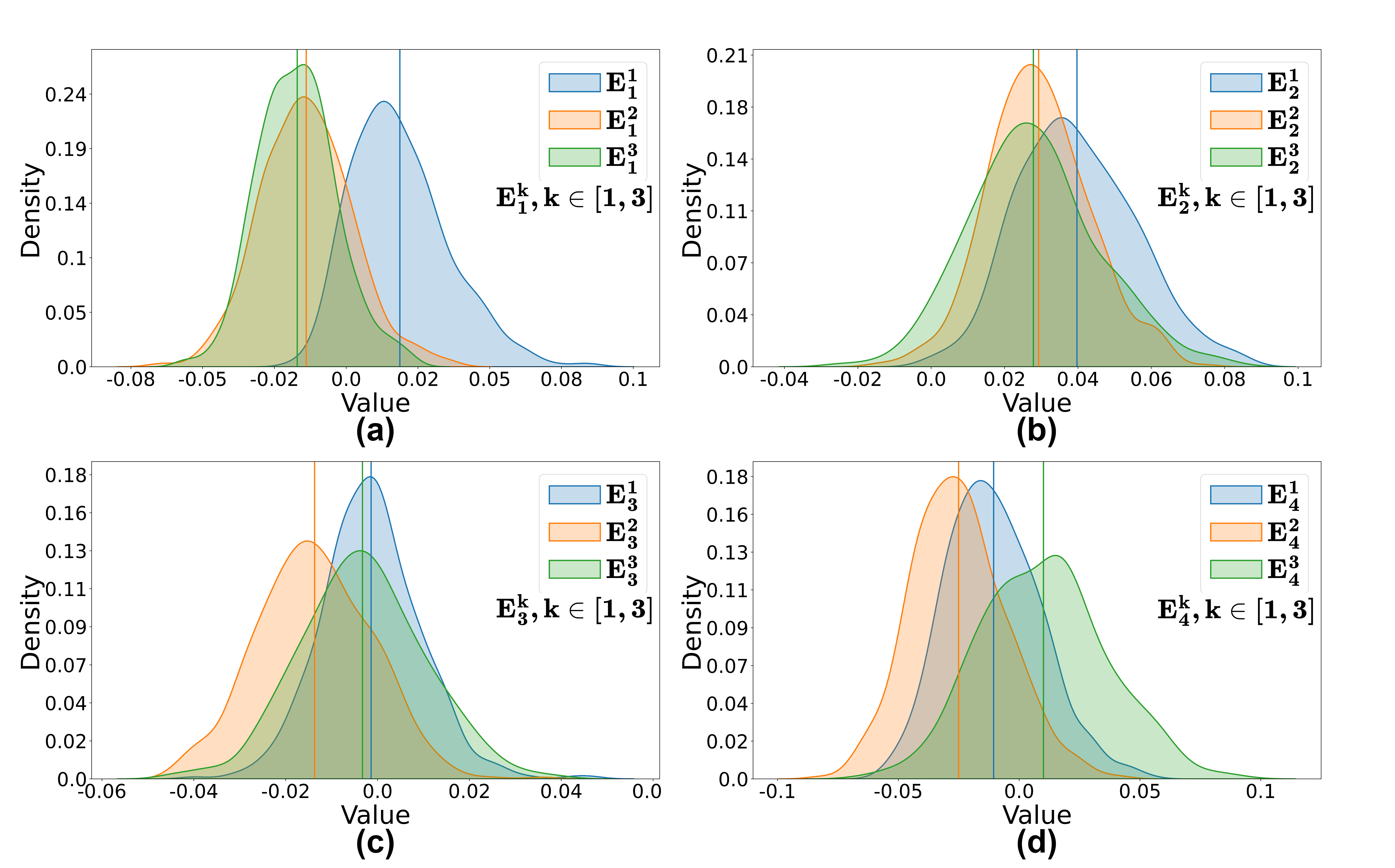}
   \caption{Distributions of the same feature across different classes in CIFAR100 training set. We examine the first 4 feature distributions for 3 classes, denoted as $\ebf^\cls_i,i\in[1,4],\cls\in[1,3]$. Panels (a) to (d) show the distributions of $\ebf^\cls_1$ to $\ebf^\cls_4$ for these three classes. The vertical lines represent the mean values of each distribution. As shown, all features approximately follow a Gaussian distribution, with different combinations of means  across different classes.}
\label{fig:distribution}
\end{figure}

We verified this idea on CIFAR100~\cite{cifar100} dataset by using the embeddings of the training images as the {\labelpool}.~The experimental results are shown in \cref{tab:image-label}.~When 30\% of the training data is used as the {\labelpool}, the classification accuracy already surpasses that achieved by using the text labels as {\labelpool}. With the entire set of training data used as the {\labelpool}, testing accuracy improves by almost 5\% compared to using text labels. Since we used CLIP as the foundation model and leveraged its powerful embedding capability, we refer to our approach as LVP-CLIP. The drawback of LVP-CLIP is that the computational and memory complexity increases as the pool size $\poolsize$ grows, which will be addressed next.

\subsection{Designing More Efficient LVP }
\label{sec:generatelabelpool}
To address the computational and memory complexity while maintaining performance, we propose three different designs using {\labelpool}: (i) \textbf{LVP-CLIP-I} uses only image embeddings in the LVP, and reduces the pool size $\poolsize$ to 1, so that its inference complexity is the same as the zero-shot learning while providing better accuracy; (ii) \textbf{LVP-CLIP-IT} 
extends the {\labelpool} by combining text embeddings with image embeddings; (iii) Unlike {\methoda} and {\methodb}, \textbf{LVP-CLIP-C} does not use the similarity function to compare the test image with LVP. It trains a classifier using information stored in LVP and applies it for classification.

The overall framework for the three variations of {\method} is shown in \cref{fig:method}.~The traditional CLIP-based classifier compares the encoded test image with the text-embedding of the class label.~This can be considered as a special case of LVP, where the label vector is the text embedding. Therefore, we also refer to it as {\method}-T.

{\bf Reducing LVP Size (\methoda).}
As noted in Sec.~\ref{sec:prelim}, using the whole training set as the {\labelpool} is effective, yet expensive in terms of computational and memory requirements.  
If we can use only one vector in the LVP, which one would be the best representative of the whole training set for each class? Each image embedding $\imgf^\cls$ is a set of image features, $\imgf^\cls = \{\ebf^\cls_1, \ebf^\cls_2,\cdots,\ebf^\cls_\ebdim \}, \text{where }\ebf^\cls\in\mathbb{R}$. We plot the distributions of those features in the same class and compare such distributions across different example classes
in \cref{fig:distribution}. In the figure, the superscript of the features indicates class index, and subscripts represent the feature index. As can be seen, within a class, each feature roughly follows a Gaussian distribution. Across classes, features have different combinations of means. Therefore, we use their \emph{mean} value as the representative label vector for each class, 
\begin{equation}
\small 
    \labelv^k = \{\bar{\Tilde{\imgf}}^k\}, \text{ and }
    \bar{\Tilde{\imgf}}^k =  \frac{1}{\Tilde{\samplenum}^k}\sum^{\Tilde{\samplenum}^k}_{i=1}{\Tilde{I}^k_i},
\label{eq:method1}
\end{equation}
where $\Tilde{I}^k_i \in \dim$ is a training image embedding of class $k\in [1,K]$ and the total number of training images from class $k$ is $\Tilde{\samplenum}^k$. As shown in \cref{tab:method}, by using the average embedding of the whole training data in each class as the {\labelpool}, we not only reduce the inference complexity but also improve the performance by about 2\% with LVP-CLIP-I.
\begin{table}
  \centering
\resizebox{0.85\linewidth}{!}{
  \begin{tabular}{cc cc cc cc cc cc}
    \toprule
     &text & 100\%  & LVP-I &LVP-IT & LVP-C\\
    \midrule
    \poolsize & 1 & 500 & 1 &1 & 1\\
    \calnum &100 &50000 &100 &100 & -\\
    Acc. & 73.3 & 78.2 & 80.1 &82.0 & 81.0\\
    \bottomrule
  \end{tabular}}
  \caption{Testing accuracy of different \method\ methods on CIFAR100. (Cosine similarity is used here.)}
  \label{tab:method}
\end{table}

{\bf Enrich Image Embedding Representation with Text (\methodb).}
The success of CLIP indicates that text embeddings and image embeddings follow similar distributions and have very high correspondence. In cases, where the distribution of training images is not similar to that  
of the testing images, information from another modality, i.e., text, may serve as an unbiased prompt. Therefore, with LVP-CLIP-IT, we design the LVP as a weighted combination of the text embedding and the average embedding $\bar{\Tilde{\imgf}}^k$ of the training images as follows:  
\begin{equation}
\small
    \labelv^k=\{\cbf^k\}, \text{ and } \cbf^k = \alpha^k \times \txtf^k + \beta^k \times \bar{\Tilde{\imgf}}^k,
\label{eq:method1}
\end{equation}
where $\alpha^k,\beta^k \in \dim,$ are trainable vectors to balance the text embedding and image embedding, respectively. Experimental results in \cref{tab:method} show that combining image embedding and text embedding improves the performance by another 2\% over LVP-I.

\subsection{LVP with a Classifier ({\methodc})}
With LVP-CLIP-I and LVP-CLIP-IT, the embedding vectors belonging to difference classes are stored, and classification is performed by comparing the distance between the embedding of a test image against each stored labeled vector using a similarity function. This process treats all feature dimensions as equally important. To handle cases, where some features should have higher importance than others in the classification process, we build a simple linear classifier that maps an embedding vector to a class prediction, $f_\theta(\cdot):\mathbb{R}^{D}\rightarrow[0,1]^K$  and train it using data only from the {\labelpool}. For a test image $\hat{\imgf}$, the prediction is:
\begin{equation}
    \text{class} = \argmax(f_\theta(\hat{I})).
\label{eq:lvpc}
\end{equation}

\subsection{Optimization Objective}
Among the three variations of LVP, {\methoda} has no trainable parameters. It solely relies on the averaging of the embeddings of training images. {\methodb} and {\methodc}, on the other hand, have a small set of parameters that are trainable.

For {\methodb}, we set $\alpha,\beta$ as task-specific parameters to avoid forgetting, and use the entropy loss to train them. Given task $t\in[1,\tasknum]$ and class $\cls^t\in[1,\clsnum^t]$, where $\tasknum$ and $\clsnum^t$ are the total number of tasks and the number of classes in task $t$, respectively, the loss function used to train $\alpha$ and $\beta$ is:
\begin{equation}
    \mathcal{L} = \mathbb{E}[-\log{\frac{\exp{(\langle \cbf^{y},\imgf\rangle)}}{\sum^{K^t}_{k^t=1}\exp{(\langle \cbf^{k^t},\imgf \rangle)}}}], t\in [1,...,\tasknum].
\label{eq:loss1}
\end{equation}
For each new task, a new set of $\alpha$ and $\beta$ will be optimized. 

The parameter $\theta$ in {\methodc} is shared among all tasks. Since it is trained using labeled vectors stored in {\labelpool}, every time a new task is added, it will be trained again with the current {\labelpool}.~The process is similar to experience replay. In this way, it also does not suffer from the forgetting. Again, entropy loss is used for the training. Given the {\labelpool} $\labelv^\cls,\cls\in[1,\clsnum]$ of each class, the loss function can be written as:
\begin{equation}
    \mathcal{L} = \mathbb{E}[-\log{\frac{\exp{(\labelv^y)}}{\sum^{K}_{k=1} \exp{(\labelv^k)}}}].
\label{eq:loss2}
\end{equation}

\subsection{Discussion on Complexity and Performance}
For all three LVP variants, the size of the {\labelpool} grows as the new tasks are learned, albeit at a very slow rate. For each class, we only need one or a few labeled vectors based on the difficulty and distribution of each dataset.~Taking ImageNet100 as an example, each labeled vector $L \in \dim$ is of size $D = 768$, which is only $0.5\%$ of the size of an image $ (3\times224\times224)$. Even for 100 classes, the overall size of {\labelpool} is $76800$,
equivalent to the size of 0.5 images. The same amount of memory is required to store the text embeddings for classes, if CLIP is used to classify input images. 


Furthermore, {\method} is task-order invariant, since each {\labelpool} is generated independently. As a result, its performance is not affected by task order. Additionally, since new label vectors do not modify the existing ones, {\method} exhibits minimal forgetting. The independence of the label vectors allows for parallel learning of different classes and making it simple to merge multiple classes —a distinct advantage over existing approaches. Finally, the LVP pool is not a rehearsal buffer, since it does not store raw images, making it less vulnerable to user privacy leakage. 
\section{Experiments}
\label{sec:exp}
We evaluate {\method} in three experiment settings: (i) class-incremental learning (CIL), (ii) domain-incremental learning (DIL), and (iii) cross-tasks incremental learning (CTIL). We compare \method\; with the SOTA methods across various categories under commensurate experimental settings. Additionally, we perform extensive ablation studies to gain deeper insights into our approach. 

{\bf Implementation details.}~We use frozen text and image encoders throughout the experiments.~For the image encoder, ViT-L/14 \cite{CLIP} is used as the backbone in all experiments.
While most CLIP-based works use cosine similarity to calculate the distance between two embeddings, we have found that L1 distance works better for image-image embeddings
and is much simpler. Therefore, we use L1 similarity for {\methoda}, and cosine similarity for {\methodb}. 
The parameter $\alpha$ in {\methodb} is initialized to 0.5 for all datasets 
while $\beta$ is consistently initialized to 1. We use SGD as the optimizer to train $\alpha$ and $\beta$, with a learning rate of 0.0001. 

For {\methodc}, the classifier is trained using labeled vectors from LVP-IT, except for datasets Core50 where no semantically unique labels are provided, and thus, LVP-I vectors are used. The classifier is trained with the ADAM optimizer, using a learning rate of 0.01. Training is stopped when the loss reaches approximately 0.1-0.05.  None of the three proposed variants of LVP-CLIP requires knowledge of the total number of classes in advance.

For the CIL experiments, we generate one label vector for each class, therefore, $P^\cls=1,\cls \in[1,...,\clsnum]$. For the DIL experiments, we generate a label vector for each domain in a class. Thus $P^\cls$ equals to the number of domains. 

An \textit{upper-bound} of classification accuracy is obtained for each experiment by assuming that all training data were available upfront and a classifier is trained based on the complete dataset by using features extracted by the CLIP image encoder. 


We use \textit{average testing accuracy} (higher is better) as our metric~\cite{NIPS2017_f8752278}. After all tasks are learned, the overall accuracy is calculated by averaging the accuracy of each task. In all the experiments except CTIL, the testing data for different tasks is of equal size, ensuring that the average test accuracy is not biased toward any specific task.

\subsection{Class Incremental Learning}
We first evaluate our method on two popular 2D image datasets namely CIFAR100~\cite{cifar100} and ImageNet100~\cite{imagenet}.
Following the setup in~\cite{Attriclip}, we select 100 classes from the original ImageNet. Both CIFAR100 and ImageNet100 are divided into 10 tasks, with each task consisting of 10 classes. 

We compare our methods with two rehearsal-based methods, iCaRL~\cite{icarl} and ARI~\cite{ARI}, and three CLIP-based methods, CoOp~\cite{CoOp}, Continual-CLIP~\cite{Continual-CLIP} and AttriCLIP~\cite{Attriclip}.
For a fair comparison, all methods are implemented using ViT-L~\cite{ViT}, except for iCaRL and ARI, which are implemented with ResNet~\cite{ResNet}.

In \cref{tab:2dcls}, the best and 2nd-best performances are shown in \textbf{bold} and with \underline{underline}, respectively. For ImageNet100, all variants of LVP-CLIP outperform other CLIP-based and experience-replay methods with significant margins. Specifically, LVP-CLIP-IT and LVP-CLIP-C provide 9.2\% and 9.3\% improvement, respectively, over the best performing CLIP-based method AttriCLIP. On CIFAR100, LVP-CLIP-IT has 0.6\% higher accuracy than AttriCLIP.
\begin{table}[h]
  \centering
  \resizebox{1\linewidth}{!}{
  \begin{tabular}{lc c c c}
    \toprule
     Method & Buffer size & CIFAR100~\cite{cifar100} & ImageNet100~\cite{imagenet}\\
    \midrule
    iCaRL~\cite{icarl} & 20/class  &49.5* &59.5*\\
    ARI~\cite{ARI} & 20/class &80.9* & 79.3*\\
    CoOp~\cite{CoOp} & 10/class &67.6* &79.3*\\
    Cont.-CLIP~\cite{Continual-CLIP} & 0 &66.7* &75.4*\\
    AttriCLIP~\cite{Attriclip} & 0 &\underline{81.4}* &83.3*\\
    \midrule
    {\methoda} &0 &80.2\phantom{*} &91.8\phantom{*}\\
    {\methodb} & 0 & \bf{82.0}\phantom{*} & \underline{92.5}\phantom{*}\\
    {\methodc} & 0 & 81.1\phantom{*} & \bf{92.6}\phantom{*}\\
    \midrule
    Upper-bound  & - & 86.5\phantom{*} & 96.0\phantom{*}\\
    \bottomrule
  \end{tabular}
  }
  \vspace{-0.2cm}
  \caption{Testing accuracy on CIFAR100 and ImageNet100. 
  Data with * is obtained from~\cite{Attriclip}.}
  \label{tab:2dcls}
  \vspace{-0.4cm}
\end{table}
%
%
%
\subsection{Domain Incremental Learning}
For these experiments, we use two popular public datasets, namely DomainNet~\cite{domainnet} and CORe50~\cite{core50}. DomainNet includes objects from 345 classes. Each object is represented by images spanning six domains: clipart, infograph, painting, quickdraw, real-world and sketch. Each domain has its own dedicated training and testing datasets. Therefore, it has 6 training tasks and 6 testing tasks.

The CORe50 dataset, on the other hand, consists of 50 classes across 11 domains. Of these, 8 domains are used for training data, presented sequentially one at a time as 8 tasks, while 3 domains are reserved for testing. Importantly, the testing domains are not part of the training process, making CORe50 also suitable as a domain adaptation dataset. 

The performance is measured as the average accuracy over all testing domains. For every class, we generate a label vector for each trained domain, resulting an LVP of size 6 for DomainNet and 8 for CORe50.

We compare our method with two regularization-based approaches, EWC~\cite{EWC} and LwF~\cite{LwF}, a rehearsal-based method, ER~\cite{ER}, and two prompt-pool-based methods, L2P~\cite{L2P} and S-Prompts~\cite{S-Prompts}. For fair comparison, all methods are implemented with ViT-L~\cite{ViT}. 
As seen in \cref{tab:dil}, for both DomainNet and CORe50 datasets, the variants of {\method} provide the top and second-best performances outperforming all the baselines. 
\begin{table}[t]
  \centering
  \resizebox{1\linewidth}{!}{
  \begin{tabular}{lc c c c c}
    \toprule
     Method & Buffer size & DomainNet~\cite{domainnet} & CORe50~\cite{core50}\\
    \midrule
    EWC~\cite{EWC} & 0  &60.0 & 75.8\\
    LwF~\cite{LwF} & 0 &61.4 &77.6\\
    ER~\cite{ER} & 50/class & 64.3 & 79.5\\
    L2P~\cite{L2P} &  0&67.6 &79.7\\
    S-Prompts~\cite{S-Prompts} & 0 &69.7 &85.2\\
    \midrule
    {\methoda} &0 &\underline{70.1} &\underline{86.1}\\
    {\methodb} & 0 & \bf{70.9} & - \\
    {\methodc} & 0 & 68.6 & \bf{89.6} \\
    \midrule
    Upper-bound  & - & 75.9 & 99.0\\
    \bottomrule
  \end{tabular}
  }
  \vspace{-0.2cm}
  \caption{Testing accuracy on DomainNet and CORe50 datasets. There is not {\methodb} result for CORe50 because the dataset does not have differentiable text labels for different classes. 
  }
  \label{tab:dil}
   \vspace{-0.4cm}
\end{table}

\begin{table*}[]
\centering
\resizebox{0.9\linewidth}{!}{
\begin{tabular}{c c c c c c c c}
\toprule
  Method  & CIFAR100 & ImageNet100 & DomainNet & CORe50 & CF100 + IN100 + DN + CR50 & Ideal & Difference\\
\hline
Tasks train/test & 10/10 &10/10 &6/6 &8/3 & 34/29 & - & -\\
\hline
L2P~\cite{L2P} &\bf{88.3} & 82.3 &67.6 &79.7 & 37.4 & 81.1 &-43.7\\
DualPrompt~\cite{DualPrompt} &\underline {86.5} &85.4 &\bf{71.8} & 84.3 &40.3 &82.9 & -42.6\\
\midrule 
\methoda    &80.2 & 91.8 & 70.1 &\underline{86.1} &\underline{80.9} &82.7 & -1.8\\
\methodb    &82.0 & \underline{92.5} & \underline{70.9} &- &\bf{81.0} &83.7 &-2.7\\
\methodc    &81.1 & \bf{92.6} & 68.6 &\bf{89.6} &79.4 &83.4 & -4.0\\
\midrule  
 Upper-bound  &86.5 &96.0 & 75.9 &99.0 &87.1 &88.9 & -1.8\\ 
\bottomrule
\end{tabular}
}
\vspace{-0.2cm}
\caption{Testing accuracy on CIFAR100 + ImageNet100 + DomainNet + CORe50. {\bf Bold} is the best and \underline{underline} is the second best.}
\label{tab:ctil}
\end{table*}

\subsection{Cross-Task Incremental Learning}
We present the Cross-Task Incremental Learning (CTIL) experimental setting, which uses an ID-UNKNOWN multi-dataset  for task-incremental learning. Here, a task refers to the unit for continual learning,
 encompassing both class-incremental and domain-incremental learning across dataset. 
Our motivation is to show how well a method can perform in a more realistic setting to continuously learn new tasks.~CTIL presents a significant challenge in continual learning, as it requires a model to perform both CIL and DIL.~We conduct experiments on a four-dataset CTIL benchmark, which combines CIFAR100, ImageNet100, DomainNet and Core50, resulting in a total of 595 (100+100+345+50) distinct object classes, divided into 34 (10+10+6+8) training tasks and 29 (10+10+6+3) testing tasks. 
The 34 training tasks are processed sequentially in a random order. 

Since this setting is too challenging for most of the existing incremental learning frameworks, we chose only L2P~\cite{L2P} and DualPrompt~\cite{DualPrompt} as the baseline methods for comparison. 
We also present the performance of our methods and the baselines on each individual dataset. In the second-to-last column of \cref{tab:ctil}, we show 
the ``Ideal" performance of each technique in the combined dataset. This is calculated as the weighted average of their accuracies on individual datasets, adjusted by the number of test samples in each. For example, the ``Ideal" score for $\text{\methoda}$ is calculated as $(80.2*10 + 91.8*10 + 70.1*6 + 86.1*3)/(10+10+6+3) = 82.7$. The \textit{Difference} between the \textit{Ideal} and the actual accuracy is shown in the last column. As a reference, the upper-bound of the classification accuracy is also provided, which is obtained by training a classifier using all the training data  with a ViT-L-based backbone.

It is not surprising that the actual accuracy is consistently lower than the ideal accuracy. This is due to two main reasons: (i) as the number of classes grows, their separation in the feature space diminishes, making it harder to distinguish them; and (ii) as tasks are trained sequentially, earlier tasks are increasingly susceptible to forgetting. Since {\methoda} does not modify previously learned knowledge, it largely avoids forgetting. The 1.8\% drop in accuracy is mainly due to more congested feature distributions. {\methodc} shows a slightly higher degradation (4.0\%), likely because its simple linear classifier does not work well in a congested feature space. Overall, the {\method}-based approaches closely approximate the ideal performance, indicating that the object classes in these four datasets remain separable in the high-dimensional feature space. This also indicates that the significant 42\% accuracy drop observed in L2P and DualPrompt is primarily attributed to forgetting. 

Although prompt-pool-based approaches show slightly better performance on certain datasets (e.g., CIFAR100 and DomainNet), they greatly suffere from forgetting as the number of classes increases. Moreover, DualPrompt and L2P require roughly twice as much computation at inference compared to {\method}. In addition, {\method}-based learning only requires forward propagation, whereas L2P and DualPrompt rely on backpropagation, making LVP-CLIP far more efficient in terms of learning complexity.

\subsection{Ablation Studies}
\label{sec:ablation}
Ablation studies are conducted to assess the effect of various design variables on performance. All studies were performed on CIFAR100.

{\bf Text Prompt Quality.}
The impact of different text prompts is shown in \cref{tab:abs3}. Here, we use {\method}-T to represent the traditional CLIP-based classification, where the text (T) embedding serves as the label vector for similarity. Two slightly different text phases are used for {\methodb} and {\method}-T and their performances are compared. As can be seen, tradiational CLIP is highly sensitive to text prompt quality, making it essential to optimize the prompt. However, the proposed {\methodb} is less susceptible to the text quality, since it uses text as supplementary information alongside image features. 
\begin{table}[h]
  \centering
  \resizebox{0.7\linewidth}{!}{
  \begin{tabular}{cc cc cc}
    \toprule
     Text & ``[cls]"  & ``a photo of a [cls]"\\
    \midrule
    LVP-CLIP-T & 65.9 &73.3\\
    \midrule
    {\methodb} & \textbf{81.3} & \bf{82.0}\\
    \bottomrule
  \end{tabular}
  }
  \vspace{-0.2cm}
  \caption{Ablation study on the impact of text prompt qualities.}
  \label{tab:abs3}
\end{table}

{\bf Initialization of $\alpha$.}
The performance of {\methodb} is sensitive to the initial value of $\alpha$ as shown in \cref{tab:abs4}. It is a good practice to set $\alpha$ to 0.5 when text quality is good, and decrease it as text quality declines. 
\begin{table}[h]
\vspace{-0.1cm}
  \centering
  \resizebox{0.85\linewidth}{!}{
  \begin{tabular}{cc cc cc cc cc cc}
    \toprule
     Initialization of $\alpha$ & 0.1  & 0.3 & 0.5 &0.7 &1.0\\
    \midrule
    {\methodb} & 80.9 & 81.6 &\bf{82.0} &81.9 &81.7\\
    \bottomrule
  \end{tabular}
  }
  \vspace{-0.2cm}
  \caption{Ablation study on the impact of $\alpha$'s initial value.}
  \label{tab:abs4}
  \vspace{-0.2cm}
\end{table}

{\bf Training dataset size.} The effect of training dataset size on {\method} is shown in \cref{tab:abs1}. Since the labeled vector is the average embedding of the training images, a larger training set generally improves the label vector’s approximation of the distribution mean, provided the data is randomly sampled. However, it appears that around 150 training images sufficient to obtain a relatively good estimation of the mean. As expected, with a small number of training images, the {\methodb} significantly outperforms {\methoda} and {\methodc}, due to the additional information provided by the text embedding. For more experimental results and analysis, please refer to the Suppl. file.
\begin{table}[h]
\vspace{-0.1cm}
  \centering
  \resizebox{0.95\linewidth}{!}{
  \begin{tabular}{cc cc cc cc cc cc cc cc cc}
    \toprule
     \# of training set &1\% &3\% &5\% & 10\% & 30\% & 50\% &100\%\\
     train data/class &5 &15 &25 &50 &150 &250 &500 \\
    \midrule
    {\methoda} &66.2 &74.9 &77.3& 78.8 & 80.0 & 79.9 & \bf{80.2}\\
    {\methodb} &74.2 &79.3 &80.3 &81.0  &81.7  &81.8  & \bf{82.0}\\
    {\methodc} &67.3 &75.5 &77.8 &79.6  &80.8  &81.1  & \bf{81.1}\\
    \bottomrule
  \end{tabular}
  }
  \vspace{-0.2cm}
  \caption{Ablation result of the \# of training images on CIFAR100.}
  \label{tab:abs1}
  \vspace{-0.3cm}
\end{table}

\section{Conclusion}
\label{sec:6}
In this paper, we have introduced a novel concept, the Label Vector Pool ({\labelpool}), which enables incremental learning without forgetting by harnessing the powerful feature extraction and encoding capabilities of CLIP. {\labelpool} blurs the distinction between batch learning and incremental learning, since it does not modify the learned model to accommodate new knowledge. This approach offers a highly cost-effective solution for incremental learning. It provides several times the memory capacity of traditional methods and superb scalability to large dataset. It significantly outperforms the SOTA approach in cross-dataset mixed class- and domain-incremental learning settings with 595 classes.

{
    \small
    \bibliographystyle{ieeenat_fullname}
    \bibliography{main}
}
\clearpage
\setcounter{page}{1}
\maketitlesupplementary

\section{Ablation study on the effect of the similarity function}
The performance of various similarity functions is compared in \cref{tab:abs2}. The LVP-CLIP-T, based only on text embeddings (which corresponds to the traditional CLIP approach), is very sensitive to the choice of the similarity function. It performs extremely poorly with L1 similarity, and the cosine similarity gives much better performance compared to L1 and L2 distances. Because of the poor match between the text-embedding and the L1 similarity, {\methodb} presents a similar outcome, i.e. gives the lowest performance with L1 similarity and favors cosine similarity. 

The {\methoda}, on the other hand, has very stable performance under different similarity functions, with L1 slightly outperforming the other two. Since L1 is also much easier to calculate than cosine similarity, we adopt it as the similarity function if the labeled vector is derived from image-embeddings. Otherwise, cosine similarity is used. 

\begin{table}[h]
  \centering
  \begin{tabular}{cc cc cc cc}
    \toprule
     Similarity functions & L1 & L2 & Cos\\
    \midrule
    LVP-CLIP-T &0.1 &65.9 &73.3\\
    \midrule
    {\methoda} & \bf{80.2} & 80.0 & 80.1\\
    \midrule
    {\methodb} & 76.8 & 81.8 & \bf{82.0}\\
    \bottomrule
  \end{tabular}
  \caption{Ablation study on the effect of the  similarity function, performed on CIFAR100 dataset.}
  \label{tab:abs2}
\end{table}

\section{Memory size of LVP}
The memory size of the label vector pool (LVP) for each dataset is shown in \cref{tab:memory}. We represent the size by using different units, including the number of floating-point numbers and the equivalent number of images with dimensions 3x224x224 pixels. The total number of floating-point numbers needed to store the LVP can be calculated as $P\times K \times D$ where $P$ is the pool size, $K$ is the total number of classes, and $D = 768$. As can be seen, even with the four datasets combined, the memory needed for LVP is equivalent to only 13.6 frames of images (or 8.2MB).

\begin{table}[h]
  \centering
  \resizebox{1\linewidth}{!}{
  \begin{tabular}{cc cc cc cc cc cc}
    \toprule
      & CF100~\cite{cifar100} &IN100~\cite{imagenet}  &DN~\cite{domainnet} &CR50~\cite{core50} &CF100+IN100+DN+CR50\\
    \midrule
    pool size P &1 &1 &6 &8 &mixed(1,1,6,8)\\
    total class K &100 &100 &345& 50 &595\\
    float number  &76,800 &76,800 &1,589,760 &307,200 & 2,050,560\\
    images  &0.5 &0.5 &10.6 &2.0 & 13.6\\
    Bytes &0.3MB &0.3MB &6.4MB &1.2MB &8.2MB\\
    \bottomrule
  \end{tabular}
  }
  \caption{Memory size required for LVP in terms of floating point numbers and the equivalent image size.}
  \label{tab:memory}
\end{table}

\section{Dataset without semantic labels}
\begin{figure*}[h]
    \centering
    \includegraphics[width=1\linewidth]{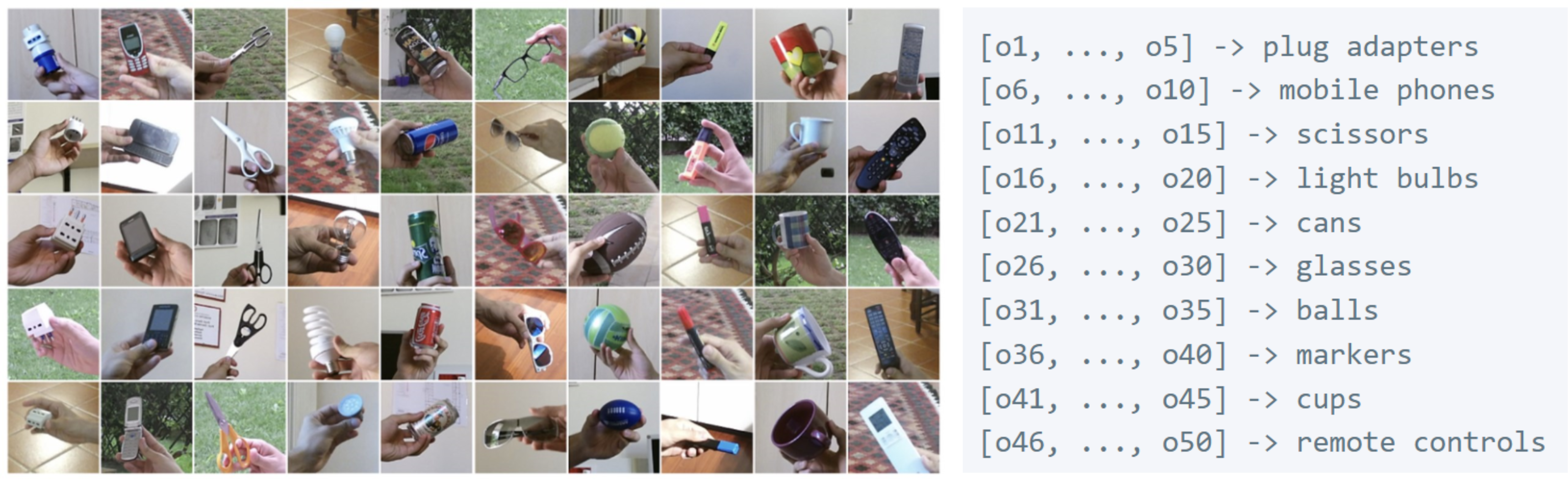}
    \caption{Images and labels from the CORe50~\cite{core50} dataset. There are a total of 50 classes but only 10 object names. Each object has five different instances as five classes. Since the class names are very close to each other as text, it is nearly impossible to separate them by zero-shot learning.}
\label{fig:core50}
\end{figure*}

As discussed in our main paper, our proposed LVP enables CLIP~\cite{CLIP}-based classification without solely relying on text embeddings. 
This is especially useful for datasets that lack meaningful text labels, such as CORe50. As shown in ~\cref{fig:core50}, CORe50 dataset has ten categories, represented by the 10 columns. Each category includes five distinct instances shown in 5 rows. Each instance is considered as an individual class. Hence, every small image in ~\cref{fig:core50} is a unique class. These classes are labeled as $o1$, $o2$, ..., $o50$ with no inherent semantic meaning. Creating a set of meaningful semantic labels to distinguish these instances is challenging. Therefore, classifying the images by comparing their features to text embeddings of the labels becomes impractical. However, with the help of the proposed LVP, the LVP-I embeddings can be easily generated from the training images, facilitating accurate classification.

\section{Unique advantages of LVP-CLIP}
\begin{figure*}[h]
    \centering
    \includegraphics[width=0.9\linewidth]{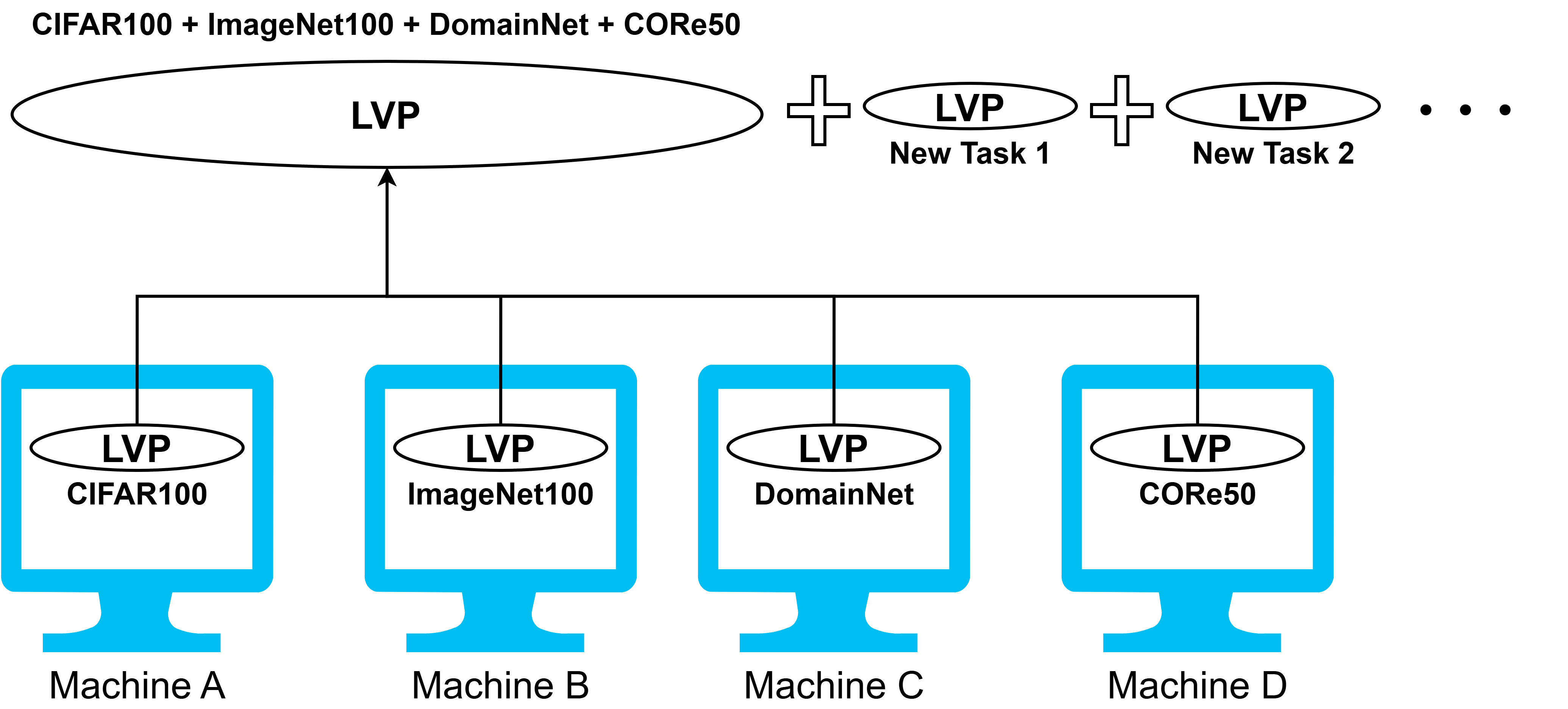}
    \caption{The illustration of the parallelizability and retaining-free continual learning ability of LVP-CLIP. Four machines conduct the experiments independently and in parallel and store the LVPs for each dataset. By simply concatenating all of LVPs, the continual learning of the four datasets is achieved. Moreover, as the new tasks arrive, the concatenation is simply repeated 
    to store the knowledge from new tasks.}
\label{fig:parallel}
\end{figure*}

As illustrated in \cref{fig:parallel}, parallel learning and retaining-free continual learning are two unique advantages of LVP-CLIP that most previous works cannot achieve. LVP-CLIP does not assume that the total number of classes is known in advance, and can learn new tasks by simply concatenating the label vector pools of each task. Moreover, since the LVPs of each task is completely independent of other tasks, the generation of LVPs can be processed on different machines in parallel.

\section{Cross-Task Incremental Learning}
\begin{figure*}[h]
    \centering
    \includegraphics[width=1\linewidth]{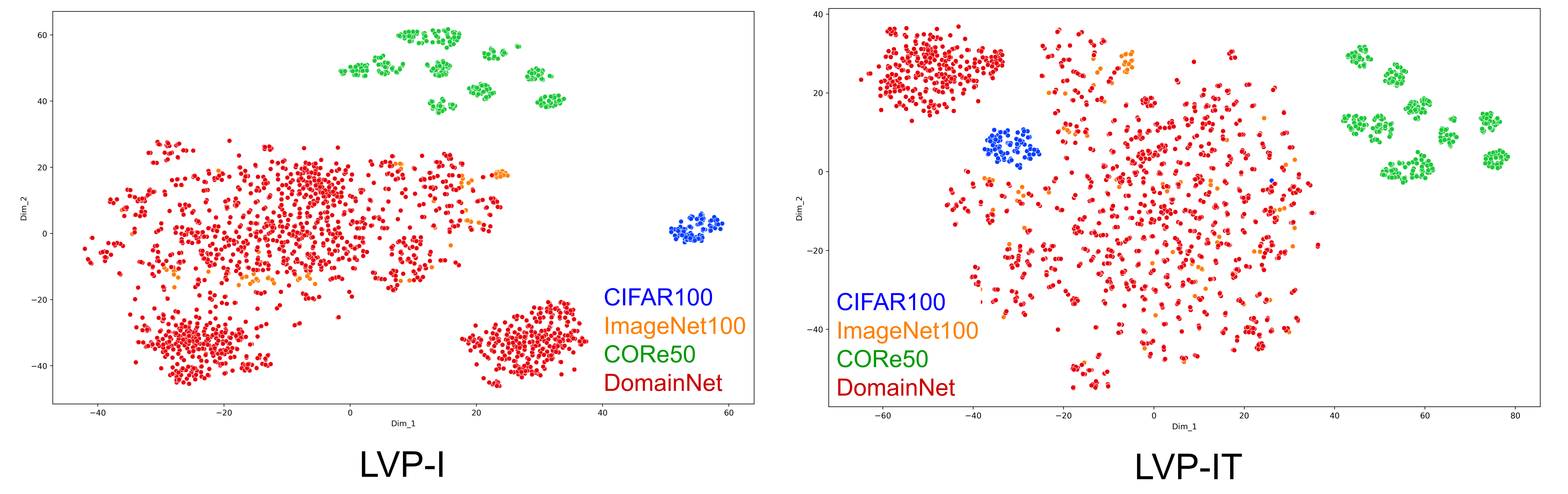}
    \caption{The T-SNE visualization of LVP-I and LVP-IT. Thanks to the remarkable feature extraction of CLIP, different dataset can be well-separated in the feature space except the ImageNet100 and DomainNet.}
\label{fig:tsne}
\end{figure*}

\cref{fig:tsne} shows the T-SNE visualization of the label vector pools generated during cross-task incremental learning (CTIL). 
As can be seen, the LVPs for different datasets are well-separated in the feature space, with the exception of ImageNet100 and DomainNet datasets. 

\cref{tab:ctil_detail} provides a detailed comparison between the ideal and actual performance of the three variants of the {\method} for each learning task. Ideal performance is defined as the test accuracy for each task when the four datasets in the CTIL setting are learned and tested independently. Entries highlighted in red indicate tasks where ideal and actual performance are closely aligned (within a difference of 0.1). As shown in \cref{fig:tsne} , the LVPs of ImageNet100 and DomainNet are intermixed, and not well separated. This explains the higher offsets observed between the Ideal and actual performances on the ten IN100 test tasks and the DN-5 task(the `real' domain) compared to the other test tasks. It is clear that, the distribution of ImageNet100 dataset is close to the `real' domain of DomainNet.

\begin{table*}[h]
\centering
\resizebox{1\linewidth}{!}{
\begin{tabular}{c c c c c c c c c c c}
\toprule
  Test Tasks  & CF100-1 & CF100-2 & CF100-3 & CF100-4 & CF100-5 & CF100-6 &CF100-7 &CF100-8 &CF100-9 &CF100-10\\
\hline
Ideal-I &{\color{RubineRed}81.9} &80.7  &{\color{RubineRed}81.1} &81.2 &{\color{RubineRed}79.3} &{\color{RubineRed}79.6} &{\color{RubineRed}74.9} &{\color{RubineRed}80.3} &{\color{RubineRed}83.8} &{\color{RubineRed}79.3}\\
Ideal-IT &85.7 &81.8  &82.0 &82.7 &81.1 &79.5 &77.1 &80.9 &87.3 &81.6\\
Ideal-C &83.1 &81.4  &82.5 &79.9 &80.2 &79.3 &75.3 &{\color{RubineRed}80.5} &86.1 &81.1\\
\midrule 
\methoda &{\color{RubineRed}82.0} & 81.0 &{\color{RubineRed}81.0} &81.0 &{\color{RubineRed}79.3} &{\color{RubineRed}79.6} &{\color{RubineRed}74.9} &{\color{RubineRed}80.3} &{\color{RubineRed}83.8} &{\color{RubineRed}79.2}\\
\methodb &85.3 &79.3 &81.1 &82.5 &80.5 &79.1 &76.5 &80.5 &85.5 &81.0\\
\methodc &84.7 &79.4 &81.7 &80.8 &76.6 &78.5 &73.6 &{\color{RubineRed}80.6} &84.5 &79.1\\

\toprule
  Test Tasks  & IN100-1 & IN100-2 & IN100-3 & IN100-4 & IN100-5 & IN100-6 &IN100-7 &IN100-8 &IN100-9 &IN100-10\\
\hline
Ideal-I &93.0 &84.2  &88.2 &97.0 &94.4 &92.0 &92.0 &90.6 &91.2 &95.4\\
Ideal-IT &93.2 &86.2  &90.2 &96.4 &94.6 &92.0 &93.6 &90.2 &92.6 &96.2\\
Ideal-C &94.4 &85.4  &90.8 &96.2 &94.8 &91.6 &93.4 &90.0 &93.6 &95.6\\
\midrule 
\methoda &89.0 &81.0 &87.2 &94.6 &84.0 &88.6 &84.0 &85.0 &86.2 &85.6\\
\methodb &90.0 &82.6 &89.4 &95.2 &85.6 &86.8 &81.2 &81.8 &83.2 &84.6\\
\methodc &90.2 &77.8 &83.6 &92.0 &81.2 &81.2 &81.2 &83.8 &80.0 &81.2\\

\toprule
  Test Tasks  & DN-1 & DN-2 & DN-3 & DN-4 & DN-5 & DN-6 &CR50-1 &CR50-2 &CR50-3 &ALL\\
\hline
Ideal-I &{\color{RubineRed}82.2} &{\color{RubineRed}56.1}  &{\color{RubineRed}76.0} &{\color{RubineRed}46.1} &87.0 &{\color{RubineRed}73.3} &{\color{RubineRed}87.0} &{\color{RubineRed}85.1} &{\color{RubineRed}86.3} &82.7\\
Ideal-IT &82.4 &58.6  &{\color{RubineRed}77.1} &{\color{RubineRed}42.1} &88.2 &74.5 &- &- &- &83.7\\
Ideal-C &80.1 &60.7  &75.5 &33.5 &87.7 &74.5 &90.3 &88.8 &{\color{RubineRed}89.6} &83.3\\
\midrule 
\methoda &{\color{RubineRed}82.2} &{\color{RubineRed}56.1} &{\color{RubineRed}75.9} &{\color{RubineRed}46.1} &86.0 &{\color{RubineRed}73.3} &{\color{RubineRed}87.0} &{\color{RubineRed}85.1} &{\color{RubineRed}86.3} &80.9\\
\methodb &81.9 & 58.2 &{\color{RubineRed}77.0} &{\color{RubineRed}42.1} &86.0 &74.2 &86.6 &84.2 &86.1 &81.0\\
\methodc &79.9 &59.8 &74.3 &31.0 &84.4 &73.9 &89.6 &87.5 &{\color{RubineRed}89.5} &79.4\\

\bottomrule
\end{tabular}
}
\caption{Results of all the cross-task incremental learning experiments. The ideal result is the test accuracy of each test task when the learning and testing are done on a given dataset independently. 
The {\method} results are the result of each test task across the four-datasets.
The numbers for which the offset from the ideal performance is less than or equal to 0.1 are highlighted in {\color{RubineRed} red} indicating nearly zero forgetting. 
}
\label{tab:ctil_detail}
\end{table*}

\section{Classes in ImageNet100}
We have selected 100 classes from ImageNet~\cite{imagenet} following \cite{Attriclip}. The label ID and class names of the 100 classes are as follows:
[15, `American robin'], [45, `Gila monster'], [54, `eastern hog-nosed snake'], [57, `garter snake'], [64, `green mamba'], [72, `European garden spider'], [90, `lorikeet'], [99, `goose'], [119, `rock crab'], [120, `fiddler crab'], [122, `American lobster'], [131, `little blue heron'], [137, `American coot'],[151, `Chihuahua'], [155, `Shih Tzu'], [157, `Papillon'], [158, `toy terrier'], [166, `Treeing Walker Coonhound'], [167, `English foxhound'], [169, `borzoi'], [176, `Saluki'], [180, `American Staffordshire Terrier'], [209, `Chesapeake Bay Retriever'], [211, `Vizsla'], [222, `Kuvasz'], [228, `Komondor'], [234, `Rottweiler'], [236, `Dobermann'], [242, `Boxer'], [246, `Great Dane'], [267, `Standard Poodle'], [268, `Mexican hairless dog [xoloitzcuintli]'], [272, `coyote'], [275, `African wild dog'], [277, `red fox'],[281, `tabby cat'],[299, `meerkat'],[305, `dung beetle'], [313, `stick insect'], [317, `leafhopper'], [331, `hare'], [342, `wild boar'], [368, `gibbon'], [374, `langur'], [407, `ambulance'], [421, `baluster handrail'],[431, `bassinet'], [449, `boathouse'], [452, `poke bonnet'], [455, `bottle cap'], [479, `car wheel'], [494, `bell or wind chime'], [498, `movie theater'], [503, `cocktail shaker'], [508, `computer keyboard'], [544, `Dutch oven'], [560, `football helmet'], [570, `gas mask or respirator'], [592, `hard disk drive'],[593, `harmonica'], [599, `honeycomb'], [606, `clothes iron'], [608, `jeans'], [619, `lampshade'],[620, `laptop computer'], [653, `milk can'], [659, `mixing bowl'], [662, `modem'], [665, `moped'], [667, `graduation cap'], [674, `mousetrap'], [682, `obelisk'],[703, `park bench'], [708, `pedestal'], [717, `pickup truck'], [724, `pirate ship'],[748, `purse'],  [758, `fishing casting reel'], [765, `rocking chair'], [766, `rotisserie'],[772, `safety pin'], [775, `sarong'], [796, `balaclava ski mask'], [798, `slide rule'],[830, `stretcher'], [854, `front curtain'], [857, `throne'], [858, `tile roof'], [872, `tripod'],[876, `hot tub'], [882, `vacuum cleaner'], [904, `window screen'], [908, `airplane wing'], [936, `cabbage'], [938, `cauliflower'], [953, `pineapple'], [959, `carbonara'],[960, `chocolate syrup'], [993, `gyromitra'], [994, `stinkhorn mushroom']]


\end{document}